\documentclass[conference]{IEEEtran}
\usepackage{cite}
\usepackage[normalem]{ulem}
\usepackage{amsmath,amssymb,amsfonts}
\usepackage[table]{xcolor}
\usepackage{algorithmic}
\usepackage{graphicx}
\usepackage{textcomp}
\usepackage{xcolor}
\usepackage{booktabs}
\usepackage[most]{tcolorbox}
\def\BibTeX{{\rm B\kern-.05em{\sc i\kern-.025em b}\kern-.08em
    T\kern-.1667em\lower.7ex\hbox{E}\kern-.125emX}}
\begin{document}

\title{DMFNet: \uline{D}ual-Backbone \uline{M}ultiscale \uline{F}usion Network for Urban Scene Classification}
\thanks{This work was conducted by Anamitra Ghosh as part of her Master's thesis at the Department of Computer Science and Engineering, Jadavpur University, Kolkata, India.}
\author{
\IEEEauthorblockN{Anamitra Ghosh  \footnote{This work was conducted by Anamitra Ghosh as part of her Master's thesis at the Department of Computer Science and Engineering, Jadavpur University, Kolkata, India.}}

\IEEEauthorblockA{
\textit{Jadavpur University}\\
anamitra108@gmail.com}
\and
\IEEEauthorblockN{Abhiroop Chatterjee}
\IEEEauthorblockA{
\textit{Jadavpur University}\\
abhiroopchat1998@gmail.com}
\and
\IEEEauthorblockN{Susmita Ghosh}
\IEEEauthorblockA{
\textit{Jadavpur University}\\
susmitaghoshju@gmail.com}
}

\maketitle

\begin{abstract}
This article presents DMFNet, a dual-backbone multiscale feature fusion framework with residual feature propagation and spatial attention for remote sensing scene classification. Existing approaches often face challenges in effectively capturing multiscale feature interactions and learning robust feature representations from complex aerial scenes with high intra-class variability and inter-class similarity. To address these limitations, the proposed framework employs two pretrained backbone networks to extract diverse hierarchical feature representations. A multiscale feature fusion mechanism with residual feature propagation is introduced to enhance feature interaction across multiple resolution levels. In addition, a spatial attention module is introduced to emphasize informative spatial regions in multi-object scenes. Further, a two-stage training strategy consisting of backbone freezing followed by selective fine-tuning is adopted to ensure stable optimization and improved generalization. Experiments conducted on the benchmark AID dataset demonstrate that the DMFNet achieves an average accuracy of 97.46\% $\pm$ 0.14\%. Ablative analysis further show the importance of various components in unison.
\end{abstract}


\section{Introduction}
Remote sensing scene classification has witnessed significant progress with the increasing availability of high-resolution aerial imagery and advances in deep learning. Early approaches relied on handcrafted feature descriptors, which were limited in capturing complex spatial patterns and semantic information. The adoption of convolutional neural networks (CNNs) enabled automatic hierarchical feature learning, while early feature fusion and multiscale frameworks demonstrated the effectiveness of multiscale representation and feature aggregation for aerial scene classification [1]–[5].

Subsequent attention-based CNN methods incorporated channel attention, multiscale attention, bidirectional fusion, feature refinement, multilevel inheritance, and adaptive transfer learning to improve discriminative capability and feature propagation [6]–[13]. More recently, transformer-based and hybrid CNN–Transformer architectures have further enhanced contextual understanding, multiscale feature interaction, and adaptive feature fusion through graph convolution, cross-attention, spatial–channel modeling, and advanced fusion strategies [14]–[29]. Despite these advances, many existing methods either rely on single-backbone architectures, limiting feature diversity, or introduce increased architectural complexity while providing limited mechanisms for effective multiscale feature propagation.

To address these limitations, we propose a dual-backbone multiscale feature fusion framework with residual feature propagation network. The proposed framework employs ConvNeXt-Tiny [30] and EfficientNet-B1 [31] as parallel backbone networks to extract diverse hierarchical feature representations. A multiscale feature fusion mechanism with residual feature propagation facilitates effective feature interaction across multiple resolutions, while spatial attention enhances discriminative feature learning by emphasizing informative regions. Furthermore, a two-stage training strategy comprising backbone freezing followed by selective fine-tuning ensures stable optimization and improved generalization. Experimental results on the AID dataset using a stratified 50:50 train–test split show an average classification accuracy of 97.46\% $\pm$ 0.14\%, indicating consistent performance.

\begin{figure}[htbp]
\centering
\includegraphics[width=\linewidth]{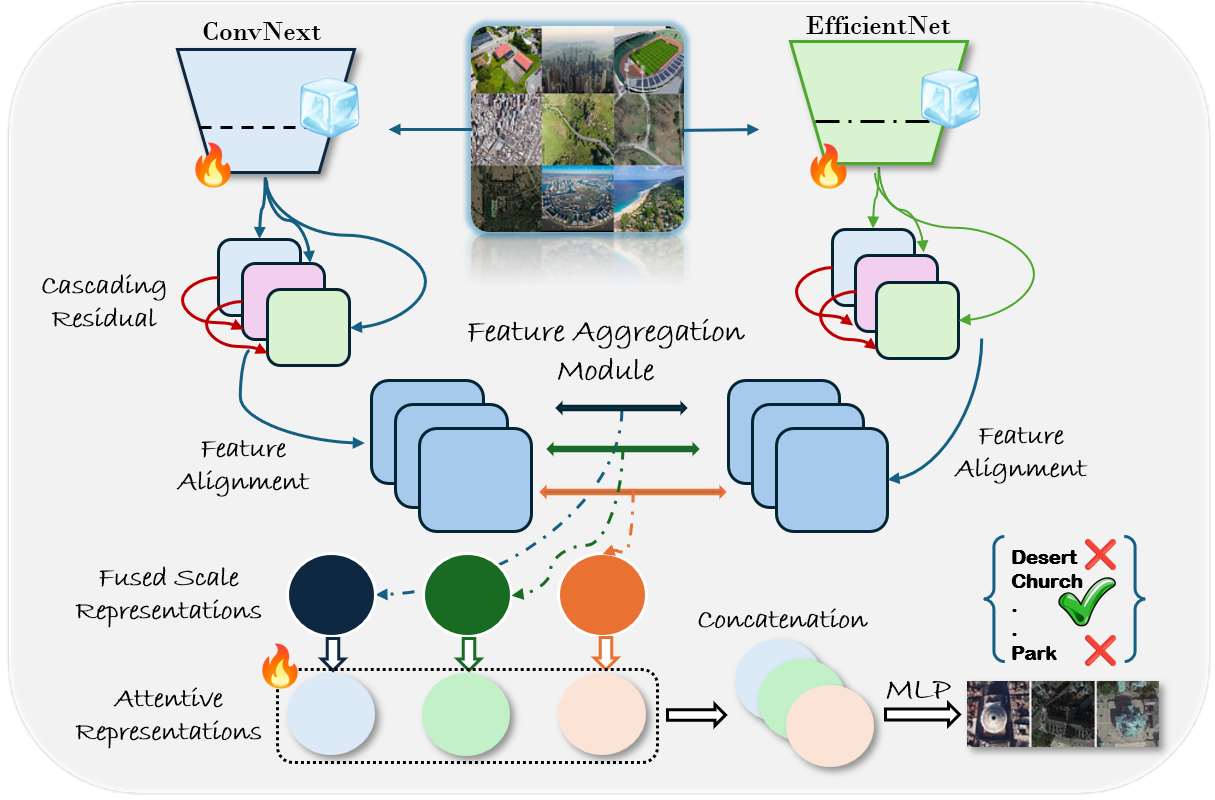}
\vspace{-0.3in}
\caption{DMFNet: Flow diagram of the proposed dual-backbone framework with multi-scale residual propagation and feature aggregation.}
\label{fig:framework}
\end{figure}

\section{Methodology}
The core idea of this work (Fig. 1) is to process input images through parallel streams to extract complementary semantic and spatial feature representations. These multi-resolution features are dynamically shared across stages via residual propagation and combined using a channel-wise fusion mechanism. A spatial attention module then refines the fused maps to emphasize distinct object boundaries while filtering out complex background noise. Finally, the network uses a robust two-stage fine-tuning strategy to generate accurate domain-adapted scene classifications. We detail the proposed approach in the following subsections comprehensively.

\subsection{Dual-Backbone Feature Extraction}
Single-backbone architectures often have limited capability to learn diverse feature representations, which can reduce their effectiveness in modeling the complex spatial arrangements and semantic variations present in remote sensing scenes. To overcome this limitation, the proposed framework employs a dual-backbone architecture comprising ConvNeXt-Tiny and EfficientNet-B1. Owing to their distinct architectural designs, the two networks learn diverse hierarchical feature representations, where ConvNeXt-Tiny captures rich semantic context while EfficientNet-B1 preserves fine-grained spatial details. The integration of these diverse representations enhances multiscale feature learning, enabling more effective modeling of complex aerial scenes.

Given an input image $X$, the two backbone networks extract hierarchical feature maps at multiple scales:
\begin{equation}
\{F_{C}^1, F_{C}^2, F_{C}^3\} = \mathcal{G}_{\text{ConvNeXt}}(X)
\end{equation}
\begin{equation}
\{F_{E}^1, F_{E}^2, F_{E}^3\} = \mathcal{G}_{\text{EfficientNet}}(X)
\end{equation}
where $\mathcal{G}_{\text{ConvNeXt}}$ and $\mathcal{G}_{\text{EfficientNet}}$ denote the feature extraction functions of ConvNeXt-Tiny and EfficientNet-B1, respectively. The feature maps $F_{C}^i$ and $F_{E}^i$ correspond to the outputs of three progressively deeper feature stages of the respective backbone networks, capturing hierarchical representations at different spatial resolutions.

Since the extracted feature maps have different channel dimensions, channel alignment is performed before multiscale feature fusion as:
\begin{equation}
\hat{F} = \text{ReLU}(\text{BN}(\text{Conv}_{1\times1}(F)))
\end{equation}
where $F$ denotes the input feature map, $\hat{F}$ is the channel-aligned feature map, and $\text{Conv}_{1\times1}$ represents the channel alignment operation. The $1\times1$ convolution projects feature maps from both backbones into a common channel space.

\subsection{Multiscale Feature Fusion with Residual Propagation}
Existing multiscale approaches often process feature maps independently at different scales, resulting in limited inter-scale interaction and reduced contextual understanding. To address this issue, a multiscale feature fusion mechanism with residual feature propagation is introduced to facilitate effective information exchange across hierarchical feature levels.

Residual feature propagation is applied to enable feature propagation across scales:
\begin{equation}
\tilde{F}_{C}^i = F_{C}^i + \mathcal{R}(F_{C}^{i-1})
\end{equation}
\begin{equation}
\tilde{F}_{E}^i = F_{E}^i + \mathcal{R}(F_{E}^{i-1})
\end{equation}
where $\mathcal{R}$ denotes spatial resizing for alignment. This residual aggregation improves gradient flow and ensures effective feature reuse across multiple scales.

Feature maps from both backbones are fused at corresponding levels using channel-wise concatenation:
\begin{equation}
F_{\text{fused}}^i = \text{Conv}_{1\times1}([\tilde{F}_{C}^i \parallel \tilde{F}_{E}^i])
\end{equation}
where $\parallel$ denotes concatenation along the channel dimension, and $\text{Conv}_{1\times1}$ represents convolution followed by batch normalization. \textit{This fusion integrates diverse feature representations to jointly capture local spatial details and high-level semantic information}.

The fused multiscale feature maps $F_{\text{fused}}^1$, $F_{\text{fused}}^2$, and $F_{\text{fused}}^3$ are further refined to $F_{\text{atten}}^i$ using spatial attention to emphasize informative spatial regions at different hierarchical levels before final multiscale aggregation.

To further integrate multiscale information, a pyramid aggregation strategy is applied:
\begin{equation}
F_{\text{final}} = \sum_{i=1}^{3} \mathcal{P}(F_{\text{atten}}^i)
\end{equation}
where $\mathcal{P}$ denotes pooling operations used for spatial alignment. This aggregation integrates information from multiple spatial resolutions to generate a robust feature representation for classification.

\subsection{Spatial Attention Mechanism}
In complex aerial scenes, a spatial attention mechanism is employed to emphasize informative regions while suppressing less relevant features.

For a feature map $F_{\text{fused}}^i$, spatial descriptors are computed using channel-wise average and max pooling:
\begin{equation}
F_{\text{avg}} = \text{AvgPool}(F_{\text{fused}}^i), \quad F_{\text{max}} = \text{MaxPool}(F_{\text{fused}}^i)
\end{equation}
These descriptors capture diverse information, where average pooling encodes global contextual information and max pooling highlights salient activations. The descriptors are concatenated and passed through a convolution operation:
\begin{equation}
M_s(F_{\text{fused}}^i) = \sigma(\text{Conv}_{7\times7}([F_{\text{avg}} \parallel F_{\text{max}}]))
\end{equation}
where $\text{Conv}_{7\times7}$ denotes a $7\times7$ convolution, and $\sigma$ is the sigmoid activation function. The resulting attention map $M_s$ represents spatial importance weights.

The refined feature map is obtained as:
\begin{equation}
F_{\text{atten}}^i = M_s(F_{\text{fused}}^i) \otimes F_{\text{fused}}^i
\end{equation}
where $\otimes$ denotes element-wise multiplication. \textit{This improves feature discrimination by emphasizing informative spatial regions}.

\section{Experiments and Results}
\subsection{Dataset Description}
Experiments were conducted on the Aerial Image Dataset (AID) [32], a widely used benchmark dataset for remote sensing scene classification. The dataset contains 10,000 high-resolution aerial images of size approximately $600\times600$ pixels categorized into 30 distinct scene classes, including residential, industrial, forest, airport, and farmland. Due to significant intra-class variability and inter-class similarity, the dataset presents considerable challenges for accurate scene classification. A stratified 50:50 train–test split is adopted to preserve balanced class distribution.

\subsection{Image Preprocessing}
Each RGB image is resized and preprocessed using EfficientNet normalization. To improve class balance and enhance model robustness, upsampling-based data augmentation is applied using random horizontal flipping, contrast adjustment, and brightness variation.

\subsection{Training Strategy and Optimization}
The model is trained with a batch size of 24 using a two-stage optimization process. In the first stage, the backbone networks are frozen, and only the newly added fusion, attention, and classification layers are trained using the Adam optimizer with a learning rate of $10^{-3}$. In the second stage, the last 40 layers of each backbone network are selectively unfrozen and fine-tuned using a reduced learning rate of $10^{-5}$.

\begin{tcolorbox}[
    enhanced,
    sharp corners,
    boxrule=0pt,
    borderline west={2.5pt}{0pt}{cyan!80!black},
    colback=gray!5!white,
    colframe=gray!5!white,
    left=6pt, right=4pt, top=5pt, bottom=5pt,
    fonttitle=\bfseries\sffamily\scriptsize,
    coltitle=cyan!90!black,
    title=THE TWO-STAGE \raisebox{-0.5em}{\includegraphics[height=1.8em]{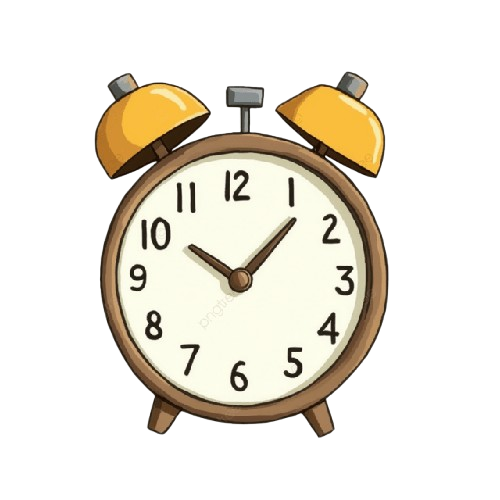}} OPTIMIZATION PIPELINE 
]
\scriptsize
\setlength{\parindent}{0pt}
Let network parameters be partitioned as $\Theta = \{\Theta_B, \Theta_H\}$, where $\Theta$ represents the total network parameters, $\Theta_B$ represents the dual backbones, and $\Theta_H$ denotes the fusion heads.

\smallskip
\textbf{Stage 1: Latent Space Alignment (Backbones Frozen \raisebox{-0.5em}{\includegraphics[height=1.8em]{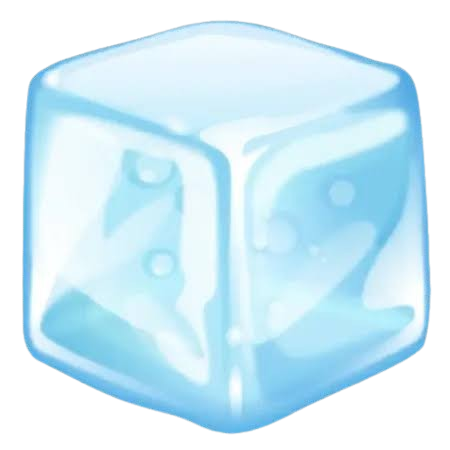}})}
$$\Theta_H^* = \arg\min_{\Theta_H} \sum_{(x,y) \in \mathcal{D}} \mathcal{L}\left( \Phi(x; \Theta_H, \Theta_B), y \right) \quad \mathbf{s.t.} \;\; \Theta_B = \Theta_B^{\text{pre}}$$
By enforcing an absolute constraint ($\mathbf{s.t.}$) locking the active backbone weights to their ImageNet-pretrained baseline ($\Theta_B^{\text{pre}}$), the search space undergoes a severe dimensional reduction. The optimization operator ($\arg\min$) isolates and updates only the head parameters ($\Theta_H$) to minimize the accumulated loss function ($\mathcal{L}$) across every input image sample $x$ and ground-truth label $y$ within the target dataset ($\mathcal{D}$). The forward mapping function ($\Phi$) computes predictions over a restricted, convex-like subspace where deep features act as deterministic base operators. This stage serves as a structural buffer, and prevents the backward propagation of unaligned gradients that would  collapse the generalized feature base to find the intermediate optimal head state ($\Theta_H^*$).

\smallskip
\textbf{Stage 2: Selective Manifold Adaptation (Joint Fine-Tuning \raisebox{-0.28em}{\includegraphics[height=1.3em]{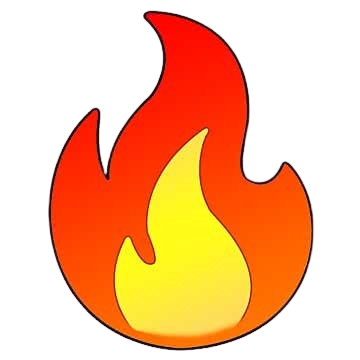}})}
$$\{\Theta_B^{(L)*}, \Theta_H^{**}\} = \arg\min_{\{\Theta_B^{(L)}, \Theta_H\}} \sum_{(x,y) \in \mathcal{D}} \mathcal{L}\left( \Phi(x; \Theta_H, \Theta_B^{(L)}), y \right)$$
$$\mathbf{s.t.} \;\; \Theta_B^{(\setminus L)} = \text{static}, \;\; \eta_{\text{stage2}} = 10^{-2} \cdot \eta_{\text{stage1}}$$
Rather than executing a risky global optimization, a selective partition unfreezes only the deepest $L = 40$ layers ($\Theta_B^{(L)}$) while early layers  ($\Theta_B^{(\setminus L)}$) capturing spatial primitives remain completely frozen. The step-down learning rate penalty ($\eta_{\text{stage2}}$) heavily scales down the initial base step-size ($\eta_{\text{stage1}}$) by a factor of $10^{-2}$ to bound the descent velocity. This acts as a highly localized manifold warping procedure that gently shifts high-level semantic boundaries to output the final converged deep backbone weights ($\Theta_B^{(L)*}$) and shifted head weights ($\Theta_H^{**}$) to adapt to the target dataset  without destabilizing foundational representations.
\end{tcolorbox}

\begin{table}[htbp]

\caption{Accuracy Comparison with State-of-the-Art Methods on AID Dataset (50:50 Train–Test Split)}
\vspace{-0.1in}
\label{tab:sota}
\centering
\small
\begin{tabular}{lccl}
\toprule
\textbf{Method} & \textbf{} & \textbf{Accuracy (\%)} \\
\midrule
\multicolumn{3}{l}{\textit{Existing Methods}} \\
\addlinespace[0.5ex]
SF-CNN [5] & \textit{TGRS '19} & $96.66 \pm 0.11$ \\
CAD [6] & \textit{JSTARS '20} & $97.16 \pm 0.26$ \\
EAM [11] & \textit{GRSL '21} & $97.06 \pm 0.19$ \\
SCViT [18] & \textit{TGRS '22} & $96.98 \pm 0.16$ \\
LG-ViT [20] & \textit{GRSL '23} & $97.67 \pm 0.15$ \\
GMFANet [23] & \textit{TGRS '24} & $\mathbf{99.73 \pm 0.09}$ \\
MSCN [26] & \textit{TGRS '25} & $97.46 \pm 0.12$ \\
SACGNet [29] & \textit{JSTARS '26} & $97.25 \pm 0.04$ \\
\midrule
\multicolumn{3}{l}{\textit{Proposed Methodology}} \\
\addlinespace[0.5ex]
\rowcolor{gray!10} \textbf{DMFNet} & \textit{} & $\mathbf{97.46 \pm 0.14}$ \\
\bottomrule
\end{tabular}
\end{table}

\subsection{Results and Analysis}
\textit{DMFNet} achieves strong and consistent classification performance on the AID dataset. Experimental evaluation over five independent runs yields an average classification accuracy of 97.46\% $\pm$ 0.14\%, demonstrating stable optimization and robust generalization. In addition, the proposed framework achieves an average precision of 97.48\% $\pm$ 0.13\%, recall of 97.46\% $\pm$ 0.14\%, and F1-score of 97.46\% $\pm$ 0.14\%, indicating balanced classification performance across diverse scene categories.

Table I compares the proposed framework with existing state-of-the-art methods on the AID dataset. Although GMFANet [23] achieves the highest reported accuracy which is attributed to there strong curriculum learning on a large pool of training dataset, the proposed framework delivers competitive performance. {In addition, it demonstrates stable learning behavior and consistent predictions across multiple experimental runs with a low variance}. This highlights its reliability for remote sensing scene classification.

\subsection{Ablative Analysis}
To evaluate the contribution of each proposed component, Table II presents the ablation study conducted on the proposed framework. M1 and M2 represent the single-backbone baselines based on ConvNeXt-Tiny and EfficientNet-B1, respectively. M3 incorporates multiscale residual feature fusion, M4 further integrates the spatial attention module, and M5 represents the complete framework with upsampling-based data augmentation. Table II confirms the effectiveness of each proposed component. Progressive integration of multiscale residual feature fusion, spatial attention, and upsampling-based data augmentation consistently improves classification accuracy, with M5 achieving the highest accuracy.

\begin{table}[htbp]
\caption{Ablation Study of the Proposed Method on the AID Dataset}
\label{tab:ablation}
\centering
\small
\begin{tabular}{ll}
\toprule
\textbf{Model} & \textbf{Accuracy (\%)} \\
\midrule
M1 (ConvNeXt-Tiny Baseline) & $95.12 \pm 0.46$ \\
M2 (EfficientNet-B1 Baseline) & $94.85 \pm 0.21$ \\
M3 (with Multiscale Fusion) & $96.91 \pm 0.16$ \\
M4 (with Spatial Attention) & $97.24 \pm 0.15$ \\
\midrule
M5 (Complete Framework) & $\mathbf{97.46 \pm 0.14}$ \\
\bottomrule
\end{tabular}
\end{table}

\section{Conclusion}
This article builds a dual-backbone multiscale feature fusion framework with residual feature propagation and spatial attention for remote sensing scene classification. By integrating ConvNeXt-Tiny and EfficientNet-B1, the proposed framework captures diverse feature representations and enhances inter-scale interaction and discriminative feature learning. Experiments on the AID dataset using a stratified 50:50 train–test split achieved an competitive average accuracy of 97.46\% $\pm$ 0.14\%, with a precision of 97.48\% $\pm$ 0.13\%, recall of 97.46\% $\pm$ 0.14\%, and F1-score of 97.46\% $\pm$ 0.14\%. Future work will focus on scaling the framework to larger datasets and enhancing computational efficiency. Specifically, we aim to explore lightweight architectures building upon our prior work [33], and investigate model compression techniques alongside the zero-shot learning paradigms in remote sensing established in [34] in the future with varying training [35] strategies.


\begin{thebibliography}{00}
\bibitem{c1} Y. Liu, Y. Liu and L. Ding, ``Scene Classification Based on Two-Stage Deep Feature Fusion,'' in \textit{IEEE Geoscience and Remote Sensing Letters}, vol. 15, no. 2, pp. 183-186, Feb. 2018.



\bibitem{c2} Y. Liu, Y. Zhong and Q. Qin, ``Scene Classification Based on Multiscale Convolutional Neural Network,'' in \textit{IEEE Transactions on Geoscience and Remote Sensing}, vol. 56, no. 12, pp. 7109-7121, Dec. 2018.
\bibitem{c3} Y. Yu and F. Liu, ``Aerial Scene Classification via Multilevel Fusion Based on Deep Convolutional Neural Networks,'' in \textit{IEEE Geoscience and Remote Sensing Letters}, vol. 15, no. 2, pp. 287-291, Feb. 2018.
\bibitem{c4} X. Lu, H. Sun and X. Zheng, ``A Feature Aggregation Convolutional Neural Network for Remote Sensing Scene Classification,'' in \textit{IEEE Transactions on Geoscience and Remote Sensing}, vol. 57, no. 10, pp. 7894-7906, Oct. 2019.
\bibitem{c5} J. Xie, N. He, L. Fang and A. Plaza, ``Scale-Free Convolutional Neural Network for Remote Sensing Scene Classification,'' in \textit{IEEE Transactions on Geoscience and Remote Sensing}, vol. 57, no. 9, pp. 6916-6928, Sept. 2019.
\bibitem{c6} W. Tong, W. Chen, W. Han, X. Li and L. Wang, ``Channel-Attention-Based DenseNet Network for Remote Sensing Image Scene Classification,'' in \textit{IEEE Journal of Selected Topics in Applied Earth Observations and Remote Sensing}, vol. 13, pp. 4121-4132, 2020.
\bibitem{c2} A. Chatterjee and S. Ghosh, ``Learning Hyperspectral Images with Curated Text Prompts for Efficient Multimodal Alignment,'' \textit{arXiv preprint arXiv:2509.22697}, Sep. 2025.


\bibitem{c8} H. Sun, S. Li, X. Zheng and X. Lu, ``Remote Sensing Scene Classification by Gated Bidirectional Network,'' in \textit{IEEE Transactions on Geoscience and Remote Sensing}, vol. 58, no. 1, pp. 82-96, Jan. 2020.
\bibitem{c9} G. Zhang et al., ``A Multiscale Attention Network for Remote Sensing Scene Images Classification,'' in \textit{IEEE Journal of Selected Topics in Applied Earth Observations and Remote Sensing}, vol. 14, pp. 9530-9545, 2021.
\bibitem{c10} H. Alhichri, A. S. Alswayed, Y. Bazi, N. Ammour and N. A. Alajlan, ``Classification of Remote Sensing Images Using EfficientNet-B3 CNN Model With Attention,'' in \textit{IEEE Access}, vol. 9, pp. 14078-14094, 2021.
\bibitem{c3} A. Chatterjee, S. Ghosh and A. Ghosh, ``Cross-Instance Contrastive Masking in Vision Transformers for Self-Supervised Hyperspectral Image Classification,'' in \textit{Workshop on Spurious Correlation and Shortcut Learning: Foundations and Solutions}, 2025.
\bibitem{c12} J. Hu, Q. Shu, J. Pan, J. Tu, Y. Zhu and M. Wang, ``MINet: Multilevel Inheritance Network-Based Aerial Scene Classification,'' in \textit{IEEE Geoscience and Remote Sensing Letters}, vol. 19, pp. 1-5, 2022.
\bibitem{c13} W. Wang, Y. Chen and P. Ghamisi, ``Transferring CNN With Adaptive Learning for Remote Sensing Scene Classification,'' in \textit{IEEE Transactions on Geoscience and Remote Sensing}, vol. 60, pp. 1-18, 2022.
\bibitem{c14} G. Wang, N. Zhang, W. Liu, H. Chen and Y. Xie, ``MFST: A Multi-Level Fusion Network for Remote Sensing Scene Classification,'' in \textit{IEEE Geoscience and Remote Sensing Letters}, vol. 19, pp. 1-5, 2022.
\bibitem{c15} K. Xu, H. Huang, P. Deng and Y. Li, ``Deep Feature Aggregation Framework Driven by Graph Convolutional Network for Scene Classification in Remote Sensing,'' in \textit{IEEE Transactions on Neural Networks and Learning Systems}, vol. 33, no. 10, pp. 5751-5765, Oct. 2022.
\bibitem{c16} X. Tang, M. Li, J. Ma, X. Zhang, F. Liu and L. Jiao, ``EMTCAL: Efficient Multiscale Transformer and Cross-Level Attention Learning for Remote Sensing Scene Classification,'' in \textit{IEEE Transactions on Geoscience and Remote Sensing}, vol. 60, pp. 1-15, 2022.
\bibitem{c17} W. Chen, S. Ouyang, W. Tong, X. Li, X. Zheng and L. Wang, ``GCSANet: A Global Context Spatial Attention Deep Learning Network for Remote Sensing Scene Classification,'' in \textit{IEEE Journal of Selected Topics in Applied Earth Observations and Remote Sensing}, vol. 15, pp. 1150-1162, 2022.
\bibitem{c18} P. Lv, W. Wu, Y. Zhong, F. Du and L. Zhang, ``SCViT: A Spatial-Channel Feature Preserving Vision Transformer for Remote Sensing Image Scene Classification,'' in \textit{IEEE Transactions on Geoscience and Remote Sensing}, vol. 60, pp. 1-12, 2022.
\bibitem{c19} Y. Yu et al., ``C²-CapsViT: Cross-Context and Cross-Scale Capsule Vision Transformers for Remote Sensing Image Scene Classification,'' in \textit{IEEE Geoscience and Remote Sensing Letters}, vol. 19, pp. 1-5, 2022.
\bibitem{c20} T. Peng, J. Yi and Y. Fang, ``A Local–Global Interactive Vision Transformer for Aerial Scene Classification,'' in \textit{IEEE Geoscience and Remote Sensing Letters}, vol. 20, pp. 1-5, 2023.
\bibitem{c21} M. Bi, M. Wang, Z. Li and D. Hong, ``Vision Transformer With Contrastive Learning for Remote Sensing Image Scene Classification,'' in \textit{IEEE Journal of Selected Topics in Applied Earth Observations and Remote Sensing}, vol. 16, pp. 738-749, 2023.
\bibitem{c22} X. Chen et al., ``Hierarchical Feature Fusion of Transformer With Patch Dilating for Remote Sensing Scene Classification,'' in \textit{IEEE Transactions on Geoscience and Remote Sensing}, vol. 61, pp. 1-16, 2023.
\bibitem{c23} Y. Zhao et al., ``Gradient-Guided Multiscale Focal Attention Network for Remote Sensing Scene Classification,'' in \textit{IEEE Transactions on Geoscience and Remote Sensing}, vol. 62, pp. 1-18, 2024.
\bibitem{c24} X. Wang, Y. Sun and P. He, ``AFIMNet: An Adaptive Feature Interaction Network for Remote Sensing Scene Classification,'' in \textit{IEEE Geoscience and Remote Sensing Letters}, vol. 22, pp. 1-5, 2025.
\bibitem{c25} X. Lu, M. Yang, Y. Chen, S. Xiong and X. Lu, ``Multibranch Fusion-Based Feature Enhance for Remote-Sensing Scene Classification,'' in \textit{IEEE Transactions on Geoscience and Remote Sensing}, vol. 63, pp. 1-17, 2025.
\bibitem{c26} J. Ma, W. Jiang, X. Tang, X. Zhang, F. Liu and L. Jiao, ``Multiscale Sparse Cross-Attention Network for Remote Sensing Scene Classification,'' in \textit{IEEE Transactions on Geoscience and Remote Sensing}, vol. 63, pp. 1-16, 2025.
\bibitem{c27} C. Shi, M. Ding and L. Wang, ``Reparameterized Feature Aggregation Convolutional Neural Network for Remote Sensing Scene Image Classification,'' in \textit{IEEE Journal of Selected Topics in Applied Earth Observations and Remote Sensing}, vol. 18, pp. 12603-12615, 2025.
\bibitem{c28} S. Wan, Z. Zhao, X. Bian and B. Li, ``Joint Network Based on Affine Transformation and Residual Connection for Remote Sensing Scene Classification,'' \textit{2025 International Conference on Advances in Electrical Engineering and Computer Applications (AEECA)}, 2025, pp. 613-618.
\bibitem{c29} C. Shi, W. Jin, Y. Shuai, M. Wu and T. Wang, ``Joint Shifted Attention and Cross-Guided Feature Fusion for Remote Sensing Scene Classification,'' in \textit{IEEE Journal of Selected Topics in Applied Earth Observations and Remote Sensing}, vol. 19, pp. 5143-5154, 2026.
\bibitem{c30} Z. Liu, H. Mao, C.-Y. Wu, C. Feichtenhofer, T. Darrell and S. Xie, ``A ConvNet for the 2020s,'' \textit{2022 IEEE/CVF Conference on Computer Vision and Pattern Recognition (CVPR)}, 2022, pp. 11966-11976.
\bibitem{c31} M. Tan and Q. Le, ``EfficientNet: Rethinking Model Scaling for Convolutional Neural Networks,'' in \textit{Proceedings of the 36th International Conference on Machine Learning (ICML)}, Long Beach, CA, USA, 2019, pp. 6105-6114.
\bibitem{c32} G. -S. Xia et al., ``AID: A Benchmark Data Set for Performance Evaluation of Aerial Scene Classification,'' in \textit{IEEE Transactions on Geoscience and Remote Sensing}, vol. 55, no. 7, pp. 3965-3981, 2017.






\bibitem{c36} A. Chatterjee, S. Ghosh, and A. Ghosh, ``Context-aware masking and learnable diffusion-guided patch refinement in transformers via sparse supervision for hyperspectral image classification,'' in \textit{Proceedings of the IEEE/CVF International Conference on Computer Vision (ICCV)}, 2025, pp. 2906-2915.

\bibitem{c33} A. Chatterjee, S. Ghosh, A. Ghosh, and E. Ientilucci, ``CASPA: Graph-Structured Concept Anchors for Modality-Agnostic Adaptation in Vision-Language Models,'' in \textit{Proceedings of the IEEE/CVF Conference on Computer Vision and Pattern Recognition (CVPR)}, 2026, pp. 31566-31576.
\bibitem{c4} Y. Huang, L. Wang, P. Zhao, Y. Zhao, Q. Yang, Y. Du and F. Ling, ``Deep Learning in Urban Green Space Extraction in Remote Sensing: A Comprehensive Systematic Review,'' in \textit{International Journal of Remote Sensing}, vol. 46, no. 3, pp. 1117-1150, 2025.

\end{thebibliography}
\end{document}